# Convolutional Neural Network for Blur Images Detection as an Alternative for Laplacian Method

Tomasz Szandała
Wroclaw University of Science and Technology
Tomasz.Szandala@pwr.edu.pl

**With the prevalence of digital cameras, the number of digital images increases quickly, which raises the demand for non-manual image quality assessment. While there are many methods considered useful for detecting blurriness, in this paper we propose and evaluate a new method that uses a deep convolutional neural network, which can determine whether an image is blurry or not. Experimental results demonstrate the effectiveness of the proposed scheme and are compared to deterministic methods using the confusion matrix.**

*Keywords: convolutional neural network, deep neural networks, image classification, computer vision, machine learning*

I. INTRODUCTION

With the popularisation of digital cameras, many home users have collected more and more digital photos. Moreover, the photographers employed for picturing ceremonies such as weddings admit that as many as 40% of the images have insufficient quality to be proposed to a client [1]. One of the key factors that lead to quality degradation is blur. Therefore, to help these users discard the photos, automatic blur detection is highly desirable, that is, to judge whether or not a given image is blurry and to determine the extent to which the image is blurry.

Blur detection is an essential and exciting task in computer vision. A vital issue in blur detection is the procedure of selecting efficient features to differentiate between distorted and clear image sections. Several methods can be employed to solve this issue, and most of them use the two-step technique to distinguish the clear and blurred regions. The first step involves handcrafting of the isolated components in an image dependent on a number of empirical data in gradient. This is followed by a binary classifier to separate the distorted and clear sections. Two essential methods of detecting blurred images are the Laplacian variance and CNN, which are the subject of this review.

There are a lot of approaches to analyse how blurry an image is, but the best and easiest one is using the variance of Laplacian method to give us a single floating point value to represent the "blurriness" of an image[2]. This method simply convolves our input image with the Laplacian operator and computes the variance. If the variance falls below a predefined threshold, we mark the image as blurry.

Conventional methodologies of edge detection such as the Laplacian of Gaussian (LOG) have an advantage of strictness. However, they need more post-processing. El-Sayed and Sennari1 indicate that edges, which are spontaneous changes in an image fragmentation can be detected via differential filters or even the LOG. The masks are firstly moved around the image, hence pixels, which are essential in offering the vital details of the edges are processed. This method of edge detection may be erroneous due to the introduction of noise in the face of mask motion around the image. CNN offers minimal operation, making it efficient in edge detection and it is reliable in limiting the effects of noise. The simulation results documented by El-Sayed and Sennari[1] demonstrate that CNN techniques achieve edge detection more efficiently than LOG because of the noise limiting advantage. Yu et al.2 report that CNN is distinct from other conventional approaches since it incorporates attribute abstraction and score extrapolation in an optimization process and generates features spontaneously from raw image. The capabilities of CNN may further be improved by substituting the multilayer perceptron (MLP) by general regression neural networks (GRNN) and support vector regression (SVR). And as reported by Yu et al.[2], CNN featuring SVR obtains best general performance, highlighting high correlation with human subjective adjudication.

Laplacian techniques on the other hand are somewhat strict in the sense that if described with the same mask, the outcome is, in most cases, similar. This makes them a bit rigid in terms of enhancing their performance. CNN, on the hand, has considerable potential for advancements. For instance, a shallow CNN feature retrieval capacity may be improved through the incorporation of SVR. However, more research is needed to enhance the functionalities of CNN in detecting blurred images.

II. LAPLACIAN METHOD

Laplacian operator is implemented to discover edges in a picture. It is, additionally, a derivative operator, but the basic contrast between different operators such as Sobel, Kirsch and Laplacian operator is that, although all the other derivatives are first order derivatives, the Laplacian operator is a second order derivative mask[3].

This method works due to the definition of the Laplacian operator itself. The Laplacian operator highlights regions of an image which contain rapid intensity changes. Just like other operators, the Laplacian is often used for edge detection. The assumption here is that if an image contains high variance, then there is a wide range of responses, both edge-like and non-edge-like, which is representative of a normal, in-focus image. But if there is very low variance, then there is a tiny range of responses, which indicates there are very few edges in the image. As we know, the more an image is blurry, the fewer edges there are.

Obviously, the challenge here is setting the correct threshold which can be quite domain dependent. Too low of a threshold and classifier will incorrectly mark images as blurry when they are not. On the other hand, too high of a threshold will not mark the images that are actually blurry as blurry.

In order to determine the correct division point, we decided to put all variances on an axis and calculate the two weight centres, one for each class. Then, we established a border between class as the weighted mean between the weight centres.

### III. CONVOLUTIONAL NEURAL NETWORKS

Convolutional neural networks (CNNs) are neural networks with sets of neurons having tied parameters[4,5,7]. Like most neural networks, they contain several filtering layers with each layer applying an affine transformation to the vector input, followed by an element-wise non-linearity. In the case of convolutional networks, the affine transformation can be implemented as a discrete convolution rather than a fully general matrix multiplication. This makes convolutional networks computationally efficient, allowing them to scale to large images. It also builds equivariance to the translation into the model (in other words, if the image is shifted by one pixel to the right, the output of the convolution is also shifted one pixel to the right; the two representations vary equally with the translation). Image-based convolutional networks typically use a pooling layer which summarises the activations of many adjacent filters with a single response. Such pooling layers may summarise the activations of groups of units with a function such as their maximum, mean, or any other filter. These pooling layers help the network be robust to small translations of the input.

CNNs have previously been used mostly for applications such as recognition of single objects in the input image. In some cases, they have been used as components of systems that solve more complicated tasks. Girshick et al. (2013) [6] use CNNs as feature extractors for a system that performs object detection and localisation.

### IV. THE EXPERIMENT

The experiment has been done on two thousand photos (fig. 1.), of different resolutions. These mostly include the author's personal photos. In order to classify them as blurry or not, an expert's knowledge has been used. While 80% of the images were used as the training dataset, the rest were used as validation collection.

Sensitivity and specificity were chosen as the comparison measurements Sensitivity (also called the true positive rate, the recall, or probability of detection in some fields) measures the proportion of actual positives that are correctly identified as such. In our case, a positive result means that the image is blurry. Specificity (also called the true negative rate) measures the proportion of actual negatives that are correctly identified as such. Here, the amount of non-blurry images are recognised as such[9].

Moreover, an accuracy metric has been measured between methods. Here, accuracy is defined as the ratio of the correctly classified instances to all the classified instances[10].

Implementation of both the methods was done in Python 3.6, using the OpenCV module for Laplacian solution and Keras/TensorFlow for the CNN approach.

#### A. Laplacian method

In the case of the code, the Laplacian method is extremely simple. The OpenCV module provides the Laplacian method that calculates the value list, for which we then compute the variance. If the variance value is above the given threshold, the image is considered not blurry, if it is below, the image is considered blurry.

*Table 1.: Confusion matrix containing the results of the classification for the Laplacian method*

|  |  | Real condition | |
|---|---|---|---|
|  |  | Is blurry | Not blurry |
| Laplacian verdict | Is blurry | 111 | 71 |
|  | Not blurry | 7 | 211 |
|  |  | **Sensitivity** | **Specificity** |
|  |  | 0.941 | 0.748 |

The results (Table 1.) show that almost 91% of the images are correctly recognised as blurry. However, one-fourth of the non-blurry images were considered blurry. Overall, the Laplacian method provides an accuracy rate of over 80%.

#### B. CNN

For the neural network parts, we used the six-layer (fig. 2.) network written in the Keras framework. The training took 30 epochs.

*Table 2.: Confusion matrix containing the results of classification for the CNN method*

|  |  | Real condition | |
|---|---|---|---|
|  |  | Is blurry | Not blurry |
| CNN verdict | Is blurry | 87 | 65 |
|  | Not blurry | 44 | 204 |
|  |  | **Sensitivity** | **Specificity** |
|  |  | 0.664 | 0.758 |

We see that the CNN method stays far behind the Laplacian method in the case of sensitivity (66% to over 90%, see tables 1 and 2); however, surprisingly, CNN has better results in specificity metric. It recognises non-blurry images with slightly better accuracy. Although the CNN classifier is 1% more specific than the Laplacian classifier, this would still not be enough to choose CNN over Laplacian.

Moreover, CNN is non-deterministic. While the Laplacian method, defined with the same mask, always gives the same verdict, CNN, trained on a different dataset or even a different seed, might results in a marginally divergent verdict.

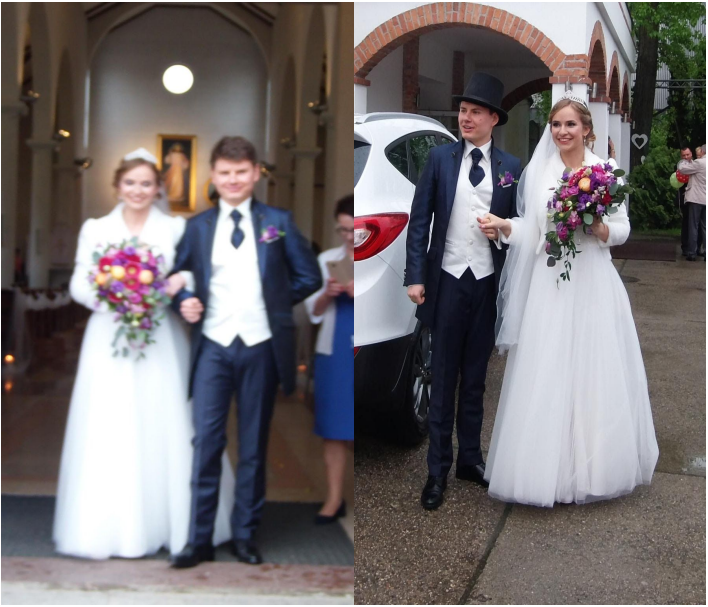

*Fig. 1: Images described by an expert as blurry (left) and sharp (right).*

## V. CONCLUSIONS

The experiment clearly shows that the Laplacian classification method is slightly superior to the CNN method.

Overall, it provides better accuracy – that is, the ratio of the correctly classified instances to all classified instances – but if we take a better look at sensitivity and specificity, more conclusions can be drawn.

The most interesting observation is that CNN has better specificity. The difference is approximately 1%, but to understand its importance we should consider its real application. If the autonomous system, used by a sample photographic service, marks a blurry image as not blurry, at the most we waste a slot in the album. On the other hand, if a non-blurry image is considered faulty and, at worst, removed from the collection, we lose an invaluable picture.

At last, this experiment proves that, presently, "deep neural networks" are not the best answer to all the problems. They are not the model answer, but only an optimisation solution for many problems.

## VI. FUTURE WORK

Not much can be done in the Laplacian method, but CNN has the potential for improvement. If we look at one of the images considered blurry by the Laplacian method and sharp by CNN, we can see that there is a section where the focus goes away (fig. 3.). This part, when compared to the rest of the picture, is too small to be meaningful for the Laplacian method. However, somehow CNN takes it into consideration. It can be investigated further, e.g., by drawing network points of interests or using Heatman on the weight activation [11].

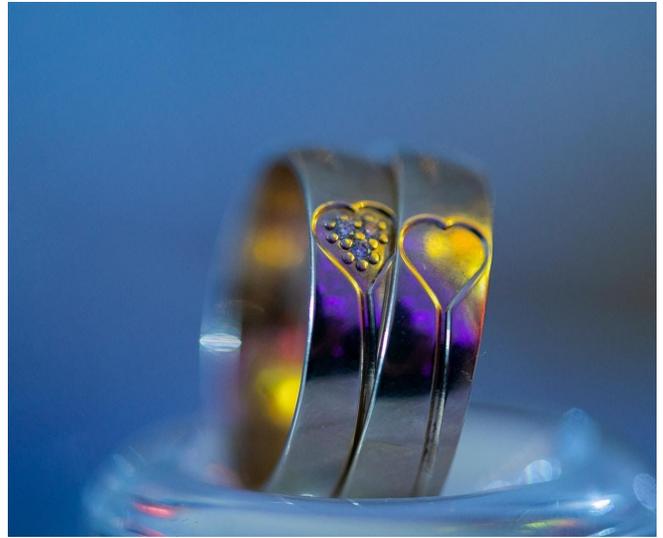

*Fig. 2.: Ambiguous image considered as non-blurry by the expert and CNN and blurry by the Laplacian method.*

## VII. REFERENCES

[1] Wang, Xin, et al. "Blind image quality assessment for measuring image blur." 2008 Congress on Image and Signal Processing. Vol. 1. IEEE (2008).

[2] Bansal, Raghav, Gaurav Raj, and Tanupriya Choudhury. "Blur image detection using Laplacian operator and Open-CV." 2016 International Conference System Modeling & Advancement in Research Trends (SMART). IEEE (2016).

[3] Wang, Xin. "Laplacian operator-based edge detectors." IEEE Transactions on Pattern Analysis and Machine Intelligence 29.5 (2007): 886–90.

[4] Fukushima, Kunihiko. "Neocognitron: A self-organizing neural network model for a mechanism of pattern recognition unaffected by shift in position." Biological cybernetics 36.4 (1980): 193–202.

[5] Ciresan, Dan Claudiu, et al. "Flexible, high performance convolutional neural networks for image classification." Twenty-Second International Joint Conference on Artificial Intelligence (2011).

[6] Girshick, Ross, and Jitendra Malik. "Training deformable part models with decorrelated features." Proceedings of the IEEE International Conference on Computer Vision (2013).

[7] Jia, Yangqing, et al. "Caffe: An open source convolutional architecture for fast feature embedding." (2013): 142.

[8] Felzenszwalb, Pedro, et al. "Visual object detection with deformable part models." Communications of the ACM 56.9 (2013): 97–105.

[9] Altman, Douglas G., and J. Martin Bland. "Diagnostic tests. 1: Sensitivity and specificity." BMJ: British Medical Journal 308.6943 (1994): 1552.

[10] Zhu, Wen, Nancy Zeng, and Ning Wang. "Sensitivity, specificity, accuracy, associated confidence interval and ROC analysis with practical SAS implementations." NESUG proceedings: health care and life sciences, Baltimore, Maryland 19 (2010): 67.

[11] Samek, Wojciech, et al. "Evaluating the visualization of what a deep neural network has learned." IEEE transactions on neural networks and learning systems 8.11 (2016): 2660–73.

[12] El-Sayed M A, Sennari H A. Convolutional Neural Network for Edge Detection in SAR Grayscale Images. Training; 12:13.

[13] Yu S, Wu S, Wang L, Jiang F, Xie Y, Li L. A shallow convolutional neural network for blind image sharpness assessment. PloS one. 2017 May 1; 12(5):e0176632.

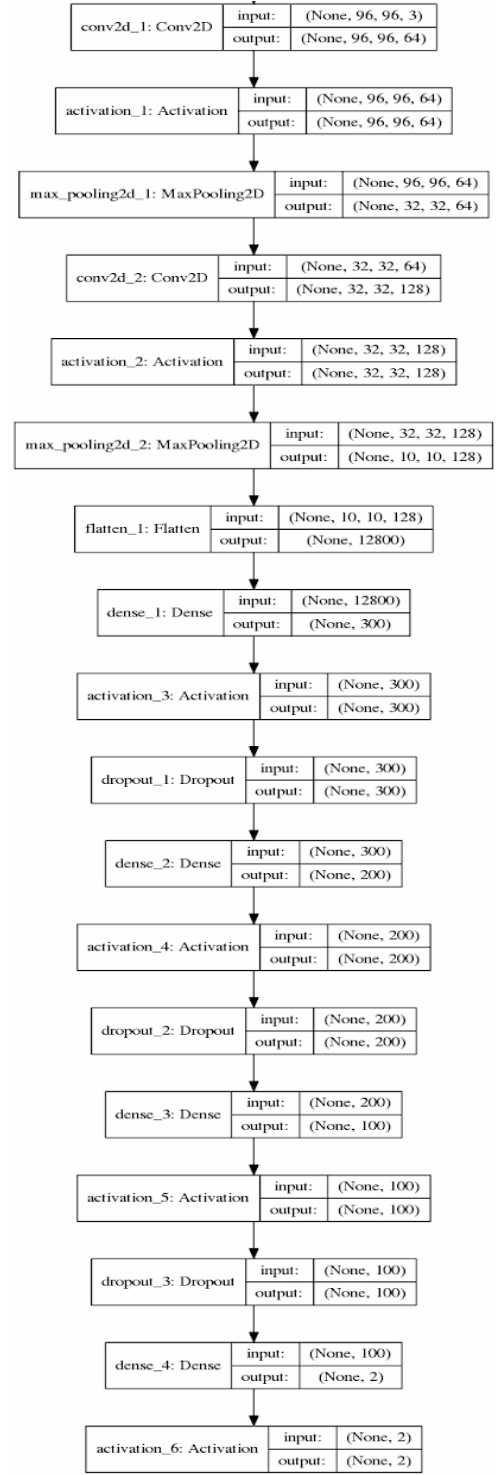

Fig 3. Schema of CNN used in the research